# Episodic Memory Deep Q-Networks

Zichuan Lin[1][3], Tianqi Zhao[2], Guangwen Yang[1], Lintao Zhang[3]

[1]Tsinghua University
[2]Microsoft
[3]Microsoft Research

linzc16@mails.tsinghua.edu.cn, tianqi.zhao@microsoft.com,
ygw@tsinghua.edu.cn, lintaoz@microsoft.com

## Abstract

Reinforcement learning (RL) algorithms have made huge progress in recent years by leveraging the power of deep neural networks (DNN). Despite the success, deep RL algorithms are known to be sample inefficient, often requiring many rounds of interaction with the environments to obtain satisfactory performance. Recently, episodic memory based RL has attracted attention due to its ability to latch on good actions quickly. In this paper, we present a simple yet effective biologically inspired RL algorithm called Episodic Memory Deep Q-Networks (EMDQN), which leverages episodic memory to supervise an agent during training. Experiments show that our proposed method can lead to better sample efficiency and is more likely to find good policies. It only requires 1/5 of the interactions of DQN to achieve many state-of-the-art performances on Atari games, significantly outperforming regular DQN and other episodic memory based RL algorithms.

## 1 Introduction

Deep neural networks has enabled significant progress in reinforcement learning research in recent years. The seminal work Deep Q-Networks (DQN) [Mnih *et al.*, 2015] successfully learns to play Atari games at or exceeding human-level performance by combining deep convolution neural network [LeCun *et al.*, 1995] and Q-learning [Watkins and Dayan, 1992]. Since then, deep reinforcement learning has achieved notable successes in a variety of tasks such as robotics control [Amarjyoti, 2017] and the game of Go [Silver *et al.*, 2016]. Unfortunately, there are still many challenges preventing RL from being applied more broadly in practice. One major problem is sample inefficiency of current deep RL algorithms. For example, it takes DQN hundreds of millions of interactions with the environment to learn a good policy and generalize to unseen states. To avoid divergence, DQN has to use a small learning rate and learns from experiences slowly.

Existing works [Pritzel *et al.*, 2017; Blundell *et al.*, 2016; Lengyel and Dayan, 2007] propose to leverage episodic control (EC) as a data-efficient approach to solve decision-making problems. The key idea is to memorize the best episodic experiences in training and replay the highly-rewarding sequences in evaluation. These methods are non-parametric since they do not depend on a parametrized value function. In these works, episodic memories are stored and updated in a lookup table during training, and are retrieved in the agent's decision making process. Table-based Episodic Control often requires very large memory footprint, and lacks generalization comparing with DNN-based RL approaches [Pritzel *et al.*, 2017; Blundell *et al.*, 2016]. Moreover, its time complexity grows dramatically as more memories are stored [Blundell *et al.*, 2016].

In this paper, we propose Episodic Memory Deep Q-Networks (EMDQN), a novel reinforcement learning algorithm which uses episodic memory to supervise an agent's training. Our work is partially inspired by human brain in decision making and motion control [Pennartz *et al.*, 2011], where two learning systems interact and compete with each other to come up with an optimal control strategy. Our approach combines the generalization strength of DQN and the fast converging property of EC by distilling the information of episodic memory into a parametric model. Experiments show that our algorithm learns good policy faster and also with less training data than other methods. Moreover, our approach provides a direct way to alleviate overestimations of Q-values, which is a common problem for Q-learning based agents with function approximators [Thrun and Schwartz, 1993; Hasselt, 2010].

We evaluate our algorithm using arcade learning environment (ALE) [Bellemare *et al.*, 2013] and show that it not only outperforms original DQN in both accuracy and training time despite being trained with 5 times fewer data frames, but also significantly outperforms other data-efficient RL algorithms using the same amount of training data.

## 2 Background

RL considers agents which learn policy by interacting with environment. An agent is facing a sequential decision making problem, where interaction with the environment takes place at discrete time steps ($t = 0, 1, ...$). We denote environment state space by $S$, the action space $A$ and the reward space $R$. At time $t$ the agent observes state $s_t \in S$, selects an action $a_t \in A$, which results in a scalar reward $r_t \in R$ and a transition to the next state $s_{t+1} \in S$. We consider infinite horizon problems with a discounted return objective $R_t = \sum_{t'=t}^{\infty} \gamma^{t'-t} r_{t'}$, where $\gamma \in [0, 1)$ is the discount factor. The goal of the agent is to find an optimal policy $\pi : S \to A$ that maximizes its expected discounted return.

## 2.1 Deep Q-Networks

A well-established technique to address the aforementioned RL task is Q-learning, where the choice of Q-function is crucial to its success. DQN successfully uses deep neural networks as Q-function to approximate the action values $Q(s, a; \theta)$, where the term $\theta$ is parameters of the network and $(s, a)$ represent a state-action pair. Two important ingredients used in DQN are target network and experience replay [Lin, 1992]. The parameters of the neural network are optimized by using stochastic gradient descent to minimize the loss

$$(r_t + \gamma \max_{a'} Q_{\bar{\theta}}(s_{t+1}, a') - Q_{\theta}(s_t, a_t))^2, \qquad (1)$$

where the term $\bar{\theta}$ represents parameters of a target network.

Many extensions have been proposed to improve DQN. Double-DQN (DDQN) [Van Hasselt *et al.*, 2016] decouples action selection and Q-function value estimation by using two Q-Networks. Prioritized experience replay [Schaul *et al.*, 2016] increases the replay probability of experiences that have a high expected learning progress measured by temporal difference errors. Dueling networks [Wang *et al.*, 2016] use two split networks to compute state function and advantage function. Optimality Tightening method (OT) [He *et al.*, 2016] combines the strength of deep Q-learning with a constrained optimization approach to tighten optimality and encourages faster reward propagation.

In general RL problems, researchers prefer to use parametric methods (e.g. DQN, A3C [Mnih *et al.*, 2016]) due to their good ability in generalizing to novel states in stochastic environment. However, it takes longer training time due to the inefficiency of existing gradient-based optimization methods (e.g. stochastic gradient descent). Agent suffers from low data efficiency because the experiences are absorbed into parameters slowly. Indeed, real-life sample efficiency restricts robotic agents from acquiring a large amount of training data in time. Therefore, inefficiency of using samples is the problem that most of the deep RL algorithms need to address.

## 2.2 Episodic Control

Near-deterministic environments are common in daily life [Blundell *et al.*, 2016]. Previous works [Lengyel and Dayan, 2007; Blundell *et al.*, 2016; Pritzel *et al.*, 2017; Gershman and Daw, 2017] propose to use episodic memory in near-deterministic environment to improve data efficiency in RL problems. [Lengyel and Dayan, 2007] suggests that episodic memory could be used to record experiences, where an agent can later imitate the sequences of actions that previously obtained high returns, and the method is referred to episodic control. Model free episodic control (MFEC) [Blundell *et al.*, 2016] is proposed to adopt episodic control to learn good policies in one-shot fashion in high-dimension games. Later, [Gershman and Daw, 2017] propose Episodic RL which uses episodic memory to construct value estimates. Specifically, episodic RL approximates action values by retrieving samples from memory and then averaging future returns. Recently, [Pritzel *et al.*, 2017] proposes neural episodic control (NEC), which makes use of differentiable neural dictionary to store slow-changing keys and fast-updating values and then retrieves useful values by context-based lookup for action selection.

However, table-based algorithms (e.g. MFEC and NEC) lack good generalization, while scalable deep RL methods (e.g. DQN, A3C) also have the problem of slow optimization. Compared with human brain, which is believed to utilize both striatum (i.e. reflex) and hippocampus (i.e. memory) in decision making [Blundell *et al.*, 2016; Pennartz *et al.*, 2011], aforementioned algorithms only rely on a single learning system. We argue that table-based episodic control and DQN are complementary to each other. We can use striatum to achieve good generalization and use hippocampus to accelerate training process via memory module and latch on good policy quickly.

## 3 Episodic Memory Deep Q-Networks

In general, episodic memory is used for direct control, but in this paper, we utilize episodic memory to accelerate the learning of DQN. We aim to address the following aspects in DQN.

1. **Slow reward propagation**. Value-bootstrapping methods, such as Q-learning, only provide updates of one-step reward or close-by multi-step rewards (as in the case of TD($\lambda$) [Singh and Sutton, 1996]), leading to low data efficiency. This can be improved by introducing monte-carlo (MC) return as learning target. However, MC return has much higher variance than TD target. Therefore, how to make good use of MC return to better propagate reward without introducing high variance is a critical problem.

2. **Single learning model**. Most of RL algorithms depend on a single learning model. Scalable deep reinforcement learning methods (e.g. DQN, A3C) simulate striatum in human brain and learn neural decision systems, while table-based methods (e.g. MFEC, NEC) simulate hippocampus in human brain and store experiences into memory system and act upon them. In this paper, we argue that both of the methods should be considered together during training to better imitate the working mechanism of human brain.

3. **Sample inefficiency**. Interacting with real environment is expensive. It takes DQN millions of interactions with the simulated environment to converge. Although the high cost of sampling can sometimes be mitigated by prioritized experience replay [Schaul *et al.*, 2016] and modeling the environments in model-based RL [Sutton, 1991; Heess *et al.*, 2015; Gu *et al.*, 2016], other mechanisms could also help direct learning by having more efficient ways to make use of samples.

We propose Episodic Memory Deep Q-Networks (EMDQN) which utilizes table-based episodic memory to accelerate agent's training. Our agent is able to rapidly latch on highly rewarding policies even though it maintains neural networks that require many steps of slow optimization for state generalization.

Our work is partly inspired by the competitive and cooperative relationship between striatum and hippocampus [Pennartz et al., 2011]. Our algorithm leverages the strength of both systems. Specifically, it uses neural network to estimate action value and provides two learning systems or targets for our agent. One system simulates striatum that provides an inference target which we denote by $S$; another system simulates hippocampus that provides a memory target which we denote by $H$. We propose a new loss function combining the two targets using appropriate weights $\alpha$ and $\beta$:

$$L = \alpha(Q_\theta - S)^2 + \beta(Q_\theta - H)^2, \quad (2)$$

where $Q_\theta$ is known as value function parametrized by $\theta$ and it acts as the decision system in our agent. As the inference target S is expected to be inferred by the agent itself, the one-step bootstrapped target is a good candidate:

$$S(s_t, a_t) = r_t + \gamma \max_{a'} Q_\theta(s_{t+1}, a'). \quad (3)$$

The memory target $H$ is defined as the best memorized return as follows:

$$H(s_t, a_t) = \max_i R_i(s_t, a_t), i \in \{1, 2, ..., E\}, \quad (4)$$

where $E$ represents the number of episodes that the agent has experienced, and $R_i(s,a)$ represents future return when taking action $a$ under state $s$ in $i$-th episode. Specifically, $H$ is a growing table, which is indexed by state-action pairs. Each transition tuple $(s, a, r)$ along the episode is cached. At the end of the episode, $H$ is updated using the transition tuples in the reverse order. In our method, $H$ is only used for training purposes. For action selection, we only consider $Q_\theta$ and select the action which yields the highest Q-value.

Similar to [Blundell et al., 2016], we maintain a state buffer for each action and use random projection technique for state compact representation. State is projected by function $\phi$ into a low-dimension vector by multiplying a Gaussian random matrix. According to Johnson-Lindenstrauss lemma, the random projection approximately preserves relative distances in original space [Johnson and Lindenstrauss, 1984]. Instead of using random projected vectors for k-nearest neighbors (KNN) search like NEC and MFEC for value estimation, we only use them for exact-matching search, which allows us to project states to lower-dimension vectors for quick table lookup. For novel state-action pairs, we add the corresponding key-value to the memory table. For existing state-action pairs, we update their values using the greater of the existing values and the future returns in current episode. Formally, we update memory table as follows:

$$H(s_t, a_t) = \begin{cases} \max\{H(s_t, a_t), R(s_t, a_t)\}, & if(s_t, a_t) \in H \\ R(s_t, a_t), & otherwise \end{cases} \quad (5)$$

where $R(s_t, a_t)$ is Monte-Carlo return in current episode. We define $\lambda = \frac{\beta}{\alpha}$ as the relative weights of $S$ and $H$. Thus, the new objective function is:

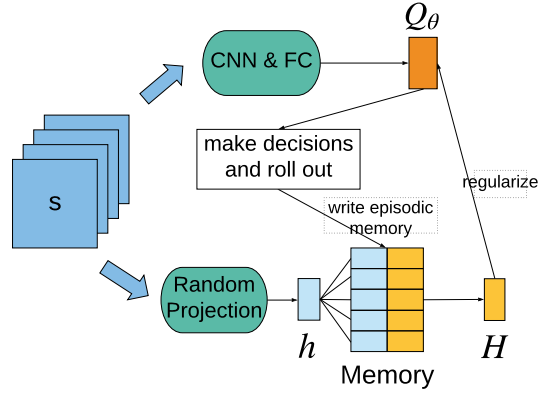

Figure 1: EMDQN architecture on a single action.

$$\min_\theta \sum_{(s_i,a_i,r_i,s_{i+1}\in D)} \Big[(Q_\theta(s_i, a_i) - S(s_i, a_i))^2 \\ + \lambda(Q_\theta(s_i, a_i) - H(s_i, a_i))^2\Big], \quad (6)$$

where $D$ represents a mini-batch of experiences.

An issue in Eq. (6) is that we cannot always find the value of $H(s_i, a_i)$ from the episodic memory table during training since some state-action pairs have not been added into $H$. To tackle this problem, we simply ignore the memory target when $(s, a)$ pairs are not found in $H$. This fits our intuition because people will never have episodic memory on things that have not happened.

**Faster reward propagation** EMDQN uses the maximum return from episodic memory to propagate rewards, compensating the disadvantage of slow-learning resulted by single step reward update. For near-deterministic environments, each reward contained in episodic memory is by far the optimal. Therefore, using memory target $H$, the rewards in the best trajectories can be propagated to parameters of $Q_\theta(s, a)$. Our algorithm differs from Eligibility Trace [Singh and Sutton, 1996] in that it keeps both bootstrapped TD target and full MC return separately rather than combining $\lambda$-return to form a new target. By learning from both targets, it quickly propagates unbiased MC return and takes advantage of low variance from TD target. Notably, episodic memory is more than unbiased MC return. It provides more concentrated learning signals and lower-variance targets than vanilla MC return since the episodic memory records the historical best MC return.

**Combination of two learning models** We combine episodic memory and DQN to better simulate the learning process of human brain. The two terms in the objective function (6) represent learning from inference target and episodic memory target respectively. We can weigh the two learning models by adjusting the value of $\lambda$: when $\lambda$ is given a small value, the method is similar to regular DQN; when given a large value, the method is closer to EC. In this way, we make flexible use of two learning methods in the learning process. The value of $\lambda$ can be adjusted higher appropriately when memory module is required and can be adjusted lower when general decision system is needed. Compared with methods that use a single model, our method is closer to the learning process of human brain. It is worth noting that EMDQN tackles generalization problem of data-efficient NEC by absorbing state features into neural networks, while NEC has many redundant state representations in its lookup table in order to find nearest neighbors.

**High sample efficiency** EMDQN introduces a mechanism to capture more information of samples. During training, it distills the best return from episodic memory and incorporates the knowledge into neural network, which makes use of the samples more efficiently. In the existing RL algorithms, all samples, regardless of the rewards, are sampled uniformly, which leads to poor training performance because non-zero rewards seldom appear. By distilling the best return of samples in every training step, our method can give more updates on non-zero reward samples. Note that our method differs from prioritized experience replay [Schaul et al., 2016]. Prioritized-DQN changes update priority of training samples. Instead, our algorithm changes the update target for each sample. This cannot be emulated by re-weighting the transitions in a prioritized-DQN. For example, a state might appear in multiple high-rewarding traces and thus getting different targets in a prioritized-DQN, while in our approach there is only a single entry in H for this state.

### 3.1 EMDQN Architecture

Figure 1 shows the EMDQN architecture. The state $s$ represented by four history frames is processed by convolution neural networks, and forward-propagated by two fully connected layers to compute $Q_\theta(s,a)$. The overall networks architecture is the same as the original DQN [Mnih et al., 2015]. State $s$ is multiplied by a random matrix drawn from Gaussian distribution and projected into a vector $h$, and passed into memory table to look up corresponding value $H(s,a)$, and then $H(s,a)$ is used to regularize $Q_\theta(s,a)$. For efficient table lookup, we use kd-tree [Bentley, 1975] to construct the memory table. All experience tuples $(\phi(s), a, r)$ along each episodic trace are cached. When updating the table, we replay each episodic trace in reverse order. Similar to the target network in DQN, we maintain a target memory table to provide stable memory value. The target memory table is updated at every K training steps using previously cached transitions. During learning, the gradients from both value-bootstrapped targets and memory targets in Eq. (6) are back-propagated to update parameter $\theta$.

## 4 Experiments

We evaluated EMDQN on the benchmark suite of 57 Atari 2600 games from the arcade learning environment [Bellemare et al., 2013]. EMDQN follows all of the networks and hyper-parameter settings as DQN as presented in [Mnih et al., 2015]. Rewards are clipped to $[-1, 1]$ when computing the true discounted return $R_t$. The coefficient $\lambda$ was tuned comparing values of $\{0.01, 0.05, 0.1, 0.2, 0.5, 1.0\}$ on the games 'Alien', 'Atlantis', 'Beamrider', 'Gopher', 'Zaxxon' but we found that larger value of $\lambda$ will deteriorate the performance. Therefore, we finally fix the value of $\lambda$ at 0.1 to regularize Q value during training. We suspect that using a dynamic gating value instead of a fixed $\lambda$ may give better performance, which we plan to investigate further in future work. For more efficient table lookup, we use random projection technique and project the states into vectors where the dimensions $dim_h$ equals to 4. Specifically, we generate a matrix with values drawn from the distribution $N(0, \frac{1}{\sqrt{dim_h}})$ and fix the matrix during training. Our state buffer size is set to 5 Million for each action, and the recent least updated state will be substituted when the buffer is full. The memory table is updated in every 10000 training steps. The experiments show that for most games episodic memory can be found with high probability (often larger than 0.95) during the training process, which leads to good regularized effect. We clip the gradient of $(Q_\theta(s_i, a_i) - S(s_i, a_i))^2$ and $(Q_\theta(s_i, a_i) - H(s_i, a_i))^2$ in Eq. (6) to $[-1, 1]$ respectively.

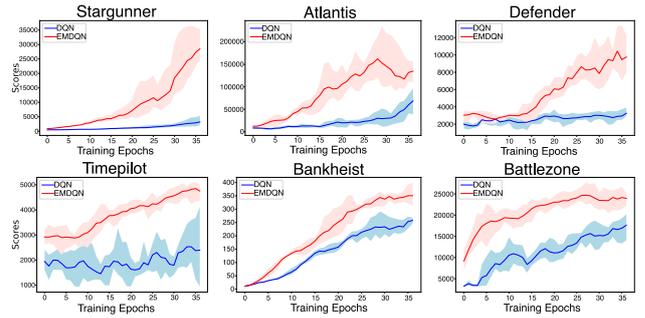

Figure 2: Testing scores for EMDQN(red), DQN(blue) on representative games. The scores are smoothed using moving average over 4 epochs. Each game is run 5 times with different random seeds.

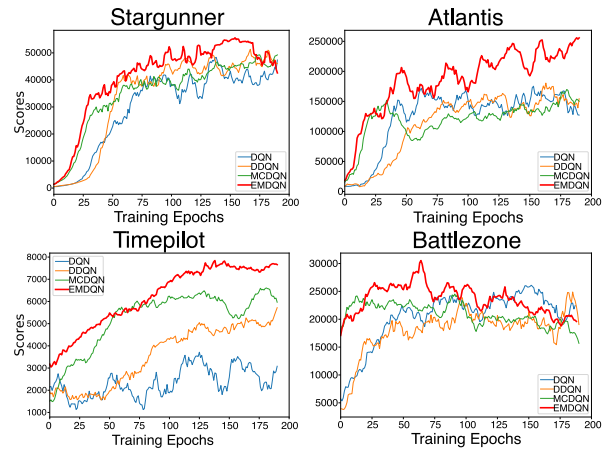

Figure 3: Training curves on 200M frames.

## 4.1 Evaluation

In the previous work of NEC [Pritzel *et al.*, 2017], different agents trained using 40 Million (40M) frames are compared. Among the seven methods (include DQN, $Q^*(\lambda)$ [Harutyunyan *et al.*, 2016], Retrace($\lambda$) [Munos *et al.*, 2016], Prioritized Replay, A3C, NEC, MFEC), the episodic memory based methods (NEC and MFEC) get state-of-the-art results. For fair comparison, we also train our EMDQN using 40M frames and compare EMDQN with NEC and MFEC to demonstrate the learning ability at the early training stage.

We train our agent for 40 epochs, each containing 1M frames. During evaluation, our agent runs for 30 episodes, each of which lasts up to 5 minutes (which corresponds to 18000 frames), and takes the averaging score as final result. Following same practice as [Blundell *et al.*, 2016], our agent starts with different initial random conditions by taking 30 no-op operations for each episode to avoid over-fitting. Following [Van Hasselt *et al.*, 2016], we use human-normalized scores to summarize the performance of our algorithm.

We show our results on 40M frames in Table 1 and the training curves of DQN and EMDQN in Figure 2. Note that our method trained on 40M frames surpasses DQN trained on 200M frames and NEC trained on 40M frames. While it is known that the learning ability of NEC will always decrease with longer training time due to the generalization problem [Pritzel *et al.*, 2017], EMDQN does not suffer from this problem. We show in Figure 3 that EMDQN can leverage the good generalization property of DQN and can keep superior learning ability in later training, covering the aforementioned shortage of NEC. We also note that the mean score of EMDQN surpasses those of all our baselines by a large margin. This is because our method performs extremely well on those games that always encounter repeated states (e.g. VideoPinball, Atlantis, Assault), which has a better match solving with episodic memory. Since our contribution is orthogonal with other techniques, we believe our approach can be combined with other DQN-relevant techniques. In addition, we test our algorithms on a variety of Atari benchmarks to demonstrate the overall performance as opposed to just cherry-picking a few examples.

We also investigate in the difference between episodic memory and monte-carlo return. To show their difference, we demonstrate an additional technique: monte carlo deep q-networks (MCDQN). MCDQN uses monte-carlo discounted future return as a substitute of episodic memory to regularize the DQN. From Figure 3, we observe that EMDQN has better learning ability than MCDQN on the games that we test. The essential difference between EMDQN and MCDQN is that EMDQN decouples the monte-carlo return from current q-networks. Instead, EMDQN stores the best monte-carlo return that the agent has seen. Therefore, we claim that the interaction shown in Figure 1 between episodic memory and DQN is crucial.

To see the effects of longer training time, we show the training curves on 200M frames on four typical games. Since the curves of NEC and MFEC are publicly unavailable, we consider the following agents: DQN, DDQN, MCDQN, EMDQN. Due to time constraint, we only run with different seeds for 40M training curves. For 200M training curves, we run all agents with the same seed. We find that EMDQN is superior to all other agents on both final results and data efficiency by a large margin. We also note that EMDQN outperforms MCDQN, showing the advantage of episodic memory over vanilla monte-carlo return.

|  | Mean | Median |
|---|---|---|
| DQN(40M) | 151.2% | 52.7% |
| MFEC(40M) | 142.2% | 61.9% |
| NEC(40M) | 144.8% | 83.3% |
| **EMDQN(40M)** | **528.4%** | **92.8%** |
| DQN(200M) | 227.9% | 79.1% |
| DDQN(200M) | 330.3% | 114.7% |

Table 1: Mean and median human-normalized scores at 40 Million frames over 57 Atari games.

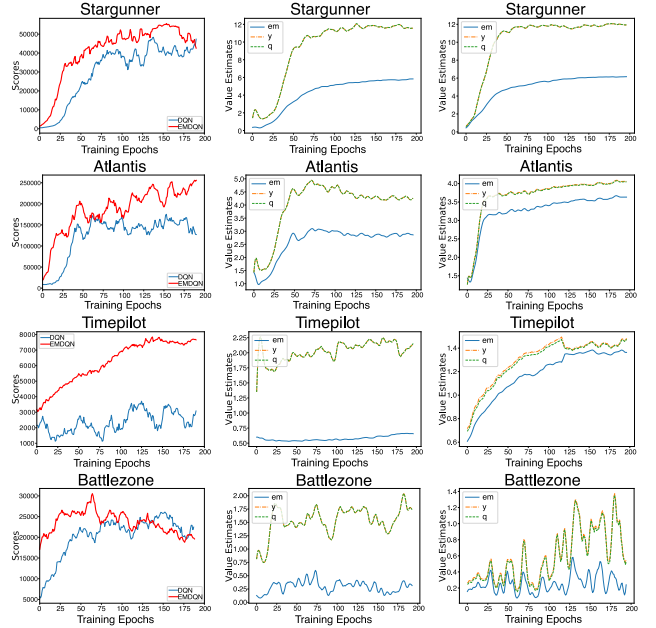

Figure 4: Left column: score curves of DQN and EMDQN; Middle column: the value of q, y and em of DQN; Right column: the value of q, y and em of EMDQN.

## 4.2 Consistent Learning Targets

To gain more constructive insights about episodic memory (EM), we intend to conduct an in-depth analysis for the training process. Since the objective function (6) consists of the Q-function, the DQN target and the EM target, we denote them by q, y and em respectively and show their value curves during training in Figure 4. In the middle column, without the constraint of episodic memory, the curves of q and y almost overlap with each other, and the q values always diverge a lot with the values of episodic memory at the beginning. One possible reason is that DQN tends to get stuck into a local minimum and learns a Q-function that goes against the actual episodic memory at the beginning. In contrast, with the constraint of episodic memory, the trends of q and episodic

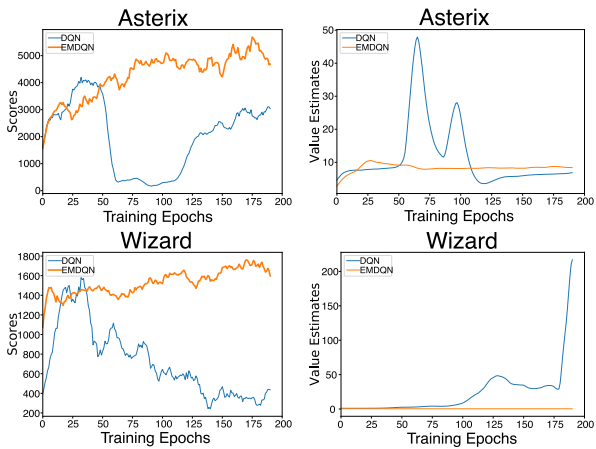

Figure 5: Left column: moving average scores; Right column: Q value during training.

memory are consistent and keep rising together.

We also find that the variance of q-function curves in EMDQN is lower than that of DQN in most cases. The agent can always learn from the best historical trajectory during training because the episodic memory records it. By doing so, the variance of learning signals is reduced. In contrast, the variance of learning signals in DQN is very high as the signals vary from sample to sample.

We find that the performance of Battlezone decreases slightly in the later training stage. We suspect that this can be made up for by reducing the value of $\lambda$ in the later training stage since a larger $\lambda$ can encourage more exploitation and less exploration. This is an interesting insight which is worth more discussion in future investigation.

### 4.3 Results on Alleviating Overoptimism

DQN is known to sometimes learn unrealistically high action quality values that lead to divergence, because the algorithm includes a maximization step over estimated action quality values [Thrun and Schwartz, 1993; Van Hasselt *et al.*, 2016].

One solution to the problem is Double-DQN [Van Hasselt *et al.*, 2016], which uses different neural networks for action selection and Q-value estimation. Although our method does not focus on alleviating the over-optimism problem, we surprisingly find that EMDQN has the potential to alleviate this problem by itself. Here we show some examples demonstrating that EMDQN can naturally alleviate overoptimism without using double Q-learning. We run both EMDQN and DQN for 200 epochs on two games and show their training curves in Figure 5. Note that in both games, the action-value of DQN blows up during training, leading to catastrophic drops of evaluation scores. In contrast, EMDQN does not suffer from overestimation and the learning is more stable. The reason is that when Q-function starts to overestimate, the episodic memory will pull down the Q-value. The degree of pulling depends on the value of $\lambda$. In this paper, we fix the value of $\lambda$ during training. However, we believe that dynamically controlling the value of $\lambda$ according to Q-value will give better results in alleviating overoptimism.

Note that the analysis here is not the major reason for the improvement of our method. The main contribution in this paper is sample-efficiency improvement in the early training stage by using episodic memory to get faster reward propagation.

## 5 Discussion

Our work is related to the optimality tightening method [He *et al.*, 2016] which uses lower and upper bounds as constraints on the value function. One might argue that the episodic memory in this paper is a special case of their upper bound. However, this is not the case. On one hand, the upper bound in [He *et al.*, 2016] is derived using backward reward, while the episodic memory target in our approach is derived from forward reward. These two are quite different. Combining the two methods could make a better use of both backward and forward reward to get more improvements. On the other hand, [He *et al.*, 2016] considers only local reward propagation, while EMDQN provides full reward propagation.

We use episodic memory aiming to improve performance of DQN in near-deterministic environment. While in stochastic environment, the episodic term here can also be considered as a regularizer. EMDQN uses $\mathrm{argmax}_a Q_\theta(s, a)$ for action selection, which will take expected Q-values into account in stochastic environment. Moreover, we can dynamically adjust the value of $\lambda$ to weigh episodic memory learning target. It will be an interesting future work to consider how to enhance the episodic estimate in stochastic environment.

It should be noted that the combination of the two learning systems is non-trivial. Instead, we are trying to build a consistent bond between these two methods. The approach proposed here, building this bond through learning targets, is simple but quite efficient. We show that this biologically inspired combination is promising in RL research field. To the best of our knowledge, EMDQN is the first work that combines these two different approaches. It will be interesting to develop other algorithms to combine a parametric approach and non-parametric approach in the future.

## 6 Conclusion

In this paper we show that our algorithm can significantly improve sample efficiency of both deep Q-learning and episodic control. We also provide an in-depth analysis on how episodic memory influences the training process of DQN. On top of major experimental outcomes, we also discover that EMDQN can also alleviate overoptimism of deep Q-learning by itself. In the future, there needs to be more research done on how to 1) dynamically tune the value of $\lambda$; 2) apply our method on stochastic environments; 3) combine parametric control and non-parametric control.